\documentclass{article}

\usepackage{arxiv}

\usepackage[utf8]{inputenc} 
\usepackage[T1]{fontenc}    
\usepackage{hyperref}       
\usepackage{url}            
\usepackage{booktabs}       
\usepackage{amsfonts}       
\usepackage{nicefrac}       
\usepackage{microtype}      
\usepackage{lipsum}
\usepackage{amsmath}
\usepackage{cleveref}
\usepackage{graphicx}
\usepackage[dvipsnames]{xcolor}

\newcommand{\neigh}[1]{\ensuremath{\mathcal{N}(#1)}}
\renewcommand{\vec}[1]{\ensuremath{\mathbf{#1}}}

\usepackage{color}
\usepackage[normalem]{ulem}

\newcommand{\dl}[1]{#1}

\title{Improving the Long-Range Performance of Gated Graph Neural Networks}

\author{
  Denis Lukovnikov \\
  Ruhr University Bochum\\
  Bochum, Germany \\
  \texttt{denis.lukovnikov@rub.de} \\
   \And
 Jens Lehmann \\
  University of Bonn\\
  Bonn, Germany \\
  \texttt{jens.lehmann@cs.uni-bonn.de} \\
   \AND
  Asja Fischer \\
  Ruhr University Bochum\\
  Bochum, Germany \\
  \texttt{asja.fischer@rub.de} \\
}

\begin{document}
\maketitle

\begin{abstract}
Many popular variants of graph neural networks (GNNs) that are capable of handling multi-relational graphs may suffer from vanishing gradients. In this work, we propose a novel GNN architecture based on the Gated Graph Neural Network with an improved ability to handle long-range dependencies in multi-relational graphs. An experimental analysis on different synthetic tasks demonstrates that the proposed architecture outperforms several popular GNN models.
\end{abstract}


\section{Introduction}
Graph Neural Networks (GNN) form a class of neural network architectures specifically designed to work with graph-structured data.
In our work, we focus on multi-relational graphs, where edges are labeled 
with different edge types.
While different GNN variants have been proposed in recent literature, to the best of our knowledge, their ability to capture long-term dependencies in graph data has not been thoroughly investigated.
Due to their local aggregation nature, many layers of a GNN must be used to capture long-range patterns (i.e., at least $K$ GNN layers are needed to incorporate any information from a node that is $K$ hops away in the graph).
However, GNNs suffer from decreasing performance when the number of layers is increased. 
\dl{Zhao et al.
\cite{pairnorm} point out that this may be due to (1) over-fitting, (2) oversmoothing, and (3) vanishing gradients.
}
Oversmoothing reffers to the phenomenon that node representations become less distinguishable from each other when more  and more layers are used in a GNN.
Even though oversmoothing (and over-squashing~\cite{alon2020bottleneck}) might compound learning difficulties with deep GNNs in general, 
problems with learning long-range 
dependencies caused by vanishing gradients 
could already be an issue in very simple graphs which should suffer less from oversmoothing. Nevertheless, while several works have investigated
over-fitting (as for example addressed by CompGCN~\cite{compgcn}) and
oversmoothing~\cite{li2018deeper,chen2019measuring,pairnorm,dropedge,yang2020revisiting}, possible vanishing gradient~\cite{hochreiter1997long,pascanu2013difficulty} problems in GNNs have so far received less attention.

\dl{The vanishing and exploding gradient problems have been studied extensively for recurrent neural networks (RNN), resulting in the development of gated update functions such as those employed in the Long Short Term Memory (LSTM~\cite{hochreiter1997long}) and the Gated Recurrent Unit (GRU~\cite{gru}), 
as well as other methods, such as gradient clipping~\cite{pascanu2013difficulty}.
Popular relational GNN architectures (such as the RGCN~\cite{rgcn}, the CompGCN~\cite{compgcn}, and the Gated GNN (GGNN)~\cite{ggnn}) may suffer from vanishing gradients when trying to learn long-range patterns in multi-relational graphs.
Using gated update functions, such as the GRU in the popular GGNN, can improve learning deep networks as it avoids vanishing gradients in the depth (i.e., vertically or between layers).
However, 
based on the way the gated functions are
currently employed in the 
GGNN architecture, it may still suffer from vanishing gradients with respect to distant nodes (i.e., horizontally) because all backpropagation paths between distant nodes consists of a series of matrix multiplications and \textsf{tanh} non-linearities.
}

In this work, we focus on improving the learning of long-range dependencies in multi-relational graphs by tackling the aforementioned vanishing gradient problem and 
developing a novel 
GGNN architecture
which we show to be advantageous in experiments on different synthetic tasks.




\section{Message Passing Networks}
\label{sec:mpnn}
Given a graph $\mathcal{G} = (\mathcal{V}, \mathcal{E})$, 
a GNN 
maps each node  $v \in \mathcal{V}$  onto  representation vectors 
$\vec{h}^{1}_v \dots \vec{h}^{Z}_v$ 
by repeatedly aggregating the the representation vectors of the immediate neighbourhoods of the nodes and updating node vectors in every 
step of the 
encoding process, each associated with one of $K$ layers of the GNN.
%
Relational GNNs must also take into account the edge types between nodes in the graph $\mathcal{G}$.
%
Following a message passing framework for GNNs (similar to~\cite{gilmer2017neural}), a single layer/step of many GNNs can be decomposed into a three-step process: 
\begin{equation}
\label{eq:gnn}
    \vec{h}_v^{(k)} = \phi ( \vec{h}_v^{(k-1)}, \gamma ( \{\mu(\vec{h}_u^{(k-1)}, \vec{h}_v^{(k-1)}, \vec{e}_{u \rightarrow v}) \}_{u \in \neigh{v}} ) )
\end{equation}
where $\mu(\cdot)$ is a function that computes a ``message'' along a graph edge, $\gamma(\cdot)$ aggregates the incoming messages into a single vector, and $\phi(\cdot)$ computes a new 
representation for node $v$.
In the equation, $\neigh{v}$ are the immediate neighbours of node $v$ in the given graph $\mathcal{G}$ and $\vec{e}_{u \rightarrow v}$ denote features associated with the edge from $u$ to $v$.
After subsequently applying Eq.~\eqref{eq:gnn} $K$ times\footnote{Optionally with different parameters for $\mu(\cdot)$, $\phi(\cdot)$ and/or $\gamma$ in every step.} on each node of the graph, the final node representations $\vec{h}_v^{(K)}$ can be used for different tasks, such as graph or node classification.

A standard message function is a simple matrix multiplication, which is used by RGCN~\cite{rgcn} and GGNN~\cite{ggnn}: 
\begin{align}
\label{eq:mu:mm}
    \mu_{\mathrm{MM}}(\vec{h}_u, e_{u \rightarrow v}) &= \vec{W}_{u \rightarrow v} \vec{h}_u \enspace ,
\end{align}
where $\vec{W}_{u \rightarrow v}$ is a $\mathbb{R}^{D \times D}$ matrix of parameters associated with edge type $e_{u \rightarrow v}$ of an edge from $u$ to $v$.

GGNNs 
 implement the update function $\phi(\cdot)$ based on Gated Recurrent Units (GRUs)~\cite{gru}:
\begin{align}
    \vec{r^{(k)}} &= \sigma (W_r \vec{\overline{h}}^{(k-1)}_v + U_r \vec{h}_v^{(k-1)} + b_r) \label{eq:ggnn:1}\\
    \vec{z^{(k)}} &= \sigma (W_z \vec{\overline{h}}^{(k-1)}_v + U_z \vec{h}_v^{(k-1)} + b_z) \label{eq:ggnn:2}\\
    \vec{\hat{h}}_v^{(k)} &= \mathrm{tanh} ( W \vec{\overline{h}}^{(k-1)}_v + U (\vec{h}^{(k-1)}_v \odot \vec{r}^{(k)}) ) \label{eq:ggnn:3}\\
    \vec{h}_v^{(k)} &= (1 - \vec{z}^{(k)}) \odot \vec{h}_v^{(k-1)} + \vec{z}^{(k)} \odot \vec{\hat{h}}_v^{(k)}  \label{eq:ggnn:4} \enspace ,
\end{align}
where $\vec{\overline{h}}^{(k-1)}_v = \gamma ( \{\mu(\vec{h}_u^{(k-1)}, \vec{h}_u^{(k-1)}, \vec{e}_{u \rightarrow v}) \}_{u \in \neigh{v}} )$ is the vector representing the aggregated neighbourhood of node $v$. A simple choice for computing $\vec{\overline{h}}^{(k-1)}_v$ would be $\sum_{u \in \neigh{v}} W_{u \rightarrow v} \vec{h}_u$, a sum of edge type-dependent linear transformations of neighbouring node vectors.

 


\section{GNN modeling and long-term dependencies}
\label{sec:method}
We develop our method starting from the GGNN~\cite{ggnn} architecture, which has been used for different NLP tasks~\cite{ggnn,bogin2019representing,beck2018graph,marcheggiani2017encoding}.
While 
the GRU used in GGNNs enables easy backpropagation over a large number of layers \textit{``vertically''} (from top-level state $\vec{h}_v^{(K)}$ of a node to its initial state $\vec{h}_v^{(0)}$),
it may 
suffer from vanishing gradients ``\emph{horizontally}'' (w.r.t. distant nodes) with increasing depth of the network.
In fact, considering the GGNN equations (Eq.~\eqref{eq:mu:mm} and Eqs.~\eqref{eq:ggnn:1}-\eqref{eq:ggnn:4}),
it becomes clear
that every backpropagation path to a distant node can suffer from vanishing or exploding gradients, similarly to a simple (non-gated) RNN, because of repeated matrix multiplications (both in $\mu(\cdot)$ and in $\phi(\cdot)=\mathrm{GRU}(\cdot)$), and $\mathrm{tanh}(\cdot)$ nonlinearities.


To achieve well-behaving gradients over a large number of hops, we make extensive use of gated functions to implement additive updates to the node states 
for integrating information from previous states as well as from neighbours (Section \ref{sec:sggnn}) and for incorporating edge type information (Section \ref{sec:message}).
We also integrate attention-based aggregation into our model (Section~\ref{sec:vgat}) and use a vector-based parameterization of relations (Section~\ref{sec:message}) to avoid overfitting and improve memory consumption.
These specific changes are elaborated in the following sections.
We call the resulting architecture the \textit{Symmetrically Gated Graph Neural Network with Relational Vectors and Graph ATtention} (SGGNN-RV-GAT).



\subsection{Symmetrically Gated Graph Neural Network}
\label{sec:sggnn}
%


For the update function $\phi(\cdot)$, following the ideas behind the LSTM, we propose to change the GRU equations such that both inputs 
of the update functions are gated similarly. 
We call the proposed update function, which is described by the following formulas, a Symmetrically Gated Recurrent Unit (SGRU):
%
\begin{align}
    &\vec{r}_h^{(k)} = \sigma (W_{r_h} \vec{\overline{h}}^{(k-1)}_v + U_{r_h} \vec{h}_v^{(k-1)} + b_{r_h}) \label{eq:sggnn:2}\\
    &\vec{r}_x^{(k)} = \sigma (W_{r_x} \vec{\overline{h}}^{(k-1)}_v + U_{r_x} \vec{h}_v^{(k-1)} + b_{r_x}) \label{eq:sggnn:2}\\
    &\vec{z}_x^{(k)} = W_{z_x} \vec{\overline{h}}^{(k-1)}_v + U_{z_x} \vec{h}_v^{(k-1)} + b_{z_x} \label{eq:sggnn:3}\\
    &\vec{z}_h^{(k)} = W_{z_h} \vec{\overline{h}}^{(k-1)}_v + U_{z_h} \vec{h}_v^{(k-1)} + b_{z_h} \label{eq:sggnn:4}\\
    &\vec{z}_u^{(k)} = W_{z_u} \vec{\overline{h}}^{(k-1)}_v + U_{z_u} \vec{h}_v^{(k-1)} + b_{z_u} \label{eq:sggnn:5}\\
    &\hat{z}_{x, i}^{(k)}, \hat{z}_{h, i}^{(k)}, \hat{z}_{u, i}^{(k)} = \mathrm{softmax} ([z_{x, i}^{(k)}, z_{h, i}^{(k)}, z_{u, i}^{(k)}]) \label{eq:sggnn:6} \\
    &\vec{\hat{h}}_v^{(k)} = \mathrm{tanh} ( W (\vec{\overline{h}}^{(k-1)}_v \odot \vec{r}_x^{(k)} ) + U (\vec{h}^{(k-1)}_v \odot \vec{r}_h^{(k)} ) ) \label{eq:sggnn:7}\\
    &\vec{h}_v^{(k)} = \vec{\hat{z}}_x^{(k)} \odot \vec{\overline{h}}^{(k-1)}_v + \vec{\hat{z}}_h^{(k)} \odot \vec{h}_v^{(k-1)} + \vec{\hat{z}}_u^{(k)} \odot \vec{\hat{h}}_v^{(k)} \label{eq:sggnn:8} \enspace .
\end{align}

The SGRU equations differ from the 
GRU equations of the GGNN (shown in \eqref{eq:ggnn:1}-\eqref{eq:ggnn:4} in the Appendix) in 
(i) introducing an additional reset gate $\vec{r}_x^{(k)}$ that is applied to the aggregated neighbour states $\vec{\overline{h}}_v^{(k-1)}$ and 
(ii) computing the output state $\vec{h}_v^{(k)}$ as a three-way mixture between the previous node state $\vec{h}_v^{(k-1)}$, the aggregated neighbour states $\vec{\overline{h}}_v^{(k-1)}$, and the candidate state $\vec{\hat{h}}_v^{(k)}$ instead of a two-way mixture between the previous node state and the candidate state.\footnote{
Note, that the mixing gates in our SGRU are similar to those in the Sentence State LSTM~\cite{zhang2018sentence}. However, the original formulation of the Sentence State LSTM is not applicable for graphs in general.}
%



\subsection{Incorporating edge type information}
\label{sec:message}

Using the simple $\mu_{\mathrm{MM}}(\cdot)$ (see Eq.~\ref{eq:mu:mm})
function for adding edge type information to a node state can be problematic because it may over-parameterize the edge type update, leading to easier over-fitting and larger parameter and memory requirements.
A possible solution to reduce the number of parameters is to use a limited number of basis matrices from which a relation matrix is composed or use other decomposition techniques to provide a more parameter-efficient edge type tensor decompositions~\cite{rgat,rgcn}.
Alternatively, CompGCN~\cite{compgcn} 
makes use of a vector-based edge type parameterization.
%
We share this idea, and 
propose the following gated message function:
\begin{align}
\label{eq:sggnn:update:1}
    &\vec{m} = \sigma (\vec{b}_{\mathrm{M}} + \vec{W}_{\mathrm{M}} \mathrm{CELU}(\vec{b}_{\mathrm{A}} + \vec{W}_{\mathrm{A}} [\vec{h}_u; \vec{a}_{u \rightarrow v}]) ) \\
    &\vec{u} = \vec{b}_{\mathrm{B}} + \vec{W}_{\mathrm{B}} \mathrm{CELU}(\vec{b}_{\mathrm{A}} + \vec{W}_{\mathrm{A}} [\vec{h}_u; \vec{a}_{u \rightarrow v}]) \\
    &\mu_{\mathrm{GCM}}(\vec{h}_u, e_{u \rightarrow v}) = 
    \vec{m} 
    \odot \vec{h}_u + (1 - \vec{m})
    \odot \vec{u}
\label{eq:sggnn:update:2}
    \enspace ,
\end{align}
where $\vec{a}_{u \rightarrow v}$ is a vector $\in \mathbb{R}^{D}$ associated with the type of the edge from node $u$ to node $v$, and $\vec{W}_\mathrm{A}$, $\vec{W}_\mathrm{B}$ and $\vec{W}_\mathrm{C}$ are trainable weight matrices that are shared between different edge types. 
The function $\mu_{\mathrm{GCM}}(\cdot)$ implements a gate that mixes (using $\vec{m}$) between the original node state $\vec{h}_u$ and the relation-aware update $\vec{u}$. 




\subsection{Attention-based Neighbourhood Aggregation}
\label{sec:vgat}


To implement the aggregation function $\gamma(\cdot)$, we adapt the scaled multi-head multiplicative attention mechanism~\cite{vaswani2017attention} for aggregation.
\dl{Attention-based neighbourhood aggregation~\cite{gat,rgat} has shown to be a useful alternative to the (scaled) sum aggregation of the GGNN or RGCN. 
}
The per-head attention distributions are computed as described for Transformers~\cite{vaswani2017attention}, with the change that the key vectors are computed using both the incoming messages $\mu(\vec{h}_u, e_{u \rightarrow v})$ as well as the edge vectors $\vec{a}_{u \rightarrow v}$. Also, we do not transform the states to obtain value vectors. 

The query and key vectors for each attention head $h$ (of in total $H$ heads), $\vec{q}_v^{(h)}$ and $\vec{k}_u^{(h)}$, are computed using head-specific linear transformations parameterized by $\vec{Q}^{(h)} \in \mathbb{R}^{D/H \times D}$ and $\vec{K}^{(h)} \in \mathbb{R}^{D/H \times 2D}$:
\begin{align}
\label{eq:keyquery}
    \vec{q}_v^{(h)} &= \vec{Q}^{(h)} \vec{h}_v \enspace , \\
    \vec{k}_u^{(h)}
    &= \vec{K}^{(h)} [\mu(\vec{h}_u, e_{u \rightarrow v}); \vec{a}_{u \rightarrow v}] \label{eq:attention:key} \enspace ,
\end{align}
where $[\cdot; \cdot]$ denotes the (vertical) concatenation of two (column) vectors and $\vec{a}_{u \rightarrow v}$ is the same edge type-specific parameter vector that is used in Eqs.~\eqref{eq:sggnn:update:1}-\eqref{eq:sggnn:update:2}.
Note that Eq.~\eqref{eq:attention:key} is slightly different from a standard attention mechanism, which would compute key vectors as $\vec{K}^{(h)} \mu(\vec{h}_u, e_{u \rightarrow v})$.

The unnormalized attention weight $w^{(h)}_{u \rightarrow v}$ for node $u$ and head $h$ is computed by a scaled dot product and normalized to attention score $\alpha^{(h)}$:
\begin{align}
    w^{(h)}_{u \rightarrow v} &= \frac{\vec{q}_v^{(h)} \cdot \vec{k}_u^{(h)}}{\sqrt{D/H}} \enspace , \\ 
    \alpha^{(h)}_{u \rightarrow v} &= \frac{\mathrm{exp}(w^{(h)}_{u \rightarrow v})}{\sum_{u^* \in \neigh{v}} \mathrm{exp}(w^{(h)}_{u^* \rightarrow v})}    \enspace .
\end{align}

Unlike multi-head attention in transformers and previous work~\cite{gat,rgat}, we do not perform linear transformations to obtain the value vectors\footnote{In our preliminary experiments, using (head-specific) linear transformations to obtain value vectors as in previous work, combined with the SGRU-based update, results in a significant decrease in  performance.}.
Instead, for every head $h$, the value vector is
obtained by splitting the vector $\mu(\vec{h}_u, e_{u \rightarrow v}) = \boldsymbol{\mu}_{u \rightarrow v} = [\boldsymbol{\mu}^{(1)}_{u \rightarrow v}, \dots, \boldsymbol{\mu}^{(H)}_{u \rightarrow v}]$ in $H$ equally sized parts (every $\boldsymbol{\mu}^{(h)}_{u \rightarrow v} \in \mathbb{R}^{D/H}$) and taking the $h$-th part.
The attention scores $\alpha^{(h)}$ and value vectors $\boldsymbol{\mu}^{(h)}_{u \rightarrow v}$ are then used to compute the neighbourhood aggregation vector $\vec{\overline{h}}^{(h)}_v$ for head $h$ 
as a weighted sum:
\begin{equation}
\vec{\overline{h}}^{(h)}_v = \sum_{u \in \neigh{v}} \alpha^{(h)}_{u \rightarrow v} \boldsymbol{\mu}^{(h)}_{u \rightarrow v} \enspace .
\end{equation} 
Then, the full neighbourhood aggregation vector $\vec{\overline{h}}_v$ is computed as a concatenation over all heads $h$:
\begin{equation}
     \vec{\overline{h}}_v = [\vec{\overline{h}}^{(1)}_v; \dots; \vec{\overline{h}}^{(H)}_v]  \enspace .
\end{equation}

\section{Experimental analysis}
In this section, we present our experiments on synthetic tasks that are aimed at assessing the ability of GNNs to model long-term dependencies in multi-relational graphs.
Experimenting on synthetic tasks with generated data allows us to evaluate specific aspects of the models in setups where we can fully control the complexity of the problem and can freely generate training examples.

We experiment with a simple sequence classification task (Section~\ref{sec:recall})
and a node classification task (Section~\ref{sec:treemax}).
Each 
experiment is presented 
in a separate subsection that
contains a description of the task and experimental results
in comparison 
to other models and of an ablation study.
We consider the following ablations of our full method (SGGNN-RV-GAT) in our experiments: (1) GGNN-RV-GAT replaces the SGRU with a normal GRU, (2) SGGNN-RV-\textit{mean} replaces the attention based aggregation function with a simple mean (equivalent to uniform attention), and (3) SGGNN-RM-GAT replaces $\mu_{\mathrm{GCM}}$ with $\mu_{\mathrm{MM}}$.
    
We compare against RGCN,
the GGNN
and a version of the RGAT described in~\cite{rgat}.
See Section~\ref{app:baselines} for more details.
\footnote{
When using the original formulations of 
these models in our experiments, we quickly ran out of memory for larger hidden dimensions of the GNN and larger numbers of layers.
Gradient accumulation significantly slows down training.
To maintain reasonable training speed for our experiments, we replace $\mu_{\mathrm{MM}}(\cdot)$ (see Eq.~\ref{eq:mu:mm}) in our RGCN and GGNN with $\mu_{\mathrm{MM-red}}(\vec{h}_u) = W_B W^*_{u \rightarrow v} W_A \vec{h}_u$, where $W^*_{u \rightarrow v}$ is an edge type specific square matrix of lower dimensionality than $W_{u \rightarrow v}$ from Eq.~\ref{eq:mu:mm} and $W_A$ and $W_B$ are matrices projecting into and out of $W^*_{u \rightarrow v}$'s dimensionality 
that
are shared for all edge types.
}

\subsection{Baselines}
\label{app:baselines}
\subsubsection{RGCN}
\label{app:rgcn}
For our RGCN~\cite{rgcn} baseline, we rely on the implementation provided by the DGL framework\footnote{\url{https://github.com/dmlc/dgl/blob/master/python/dgl/nn/pytorch/conv/relgraphconv.py}}.
We experimented with weight sharing between RGCN layers and found it to perform better.


\subsubsection{GGNN}
\label{app:ggnn}
We write our own implementation of the GGNN, basing our implementation on the code provided by the DGL framework\footnote{\url{https://github.com/dmlc/dgl/blob/master/python/dgl/nn/pytorch/conv/gatedgraphconv.py}}.
\dl{We use variational dropout on the GRU.}

\subsubsection{RGAT}
\label{app:rmgat}
We adapt ~\cite{rgat} for our attention-based baseline (RGAT).
Like in \cite{rgat}, we use relation-specific transformation matrices $\vec{W}^{(r)}$ and relation and head-specific query and key matrices $\vec{Q}^{(r,h)}$ and $\vec{K}^{(r,h)}$.

First, we define relation-dependent representations for a node, which is computed based on its current state $\vec{z}^{(k-1)}$:
\begin{equation}
    \vec{g}^{(r)}_i = \vec{W}^{(r)} \vec{z}^{(k-1)}_i \enspace .
\end{equation}
Subsequently, for every head $h$, we define the relation-specific query, key and value projections as:
\begin{align}
    \vec{q}^{(h, r)}_i &= \vec{Q}^{(h, r)} \vec{g}^{(r)}_i \\
    \vec{k}^{(h, r)}_i &= \vec{K}^{(h, r)} \vec{g}^{(r)}_i \\
    \vec{v}^{(h, r)}_i &= \vec{V}^{(h, r)} \vec{z}^{(k-1)}_i
\end{align}

We compute attention between messages $\boldsymbol{\mu}_x$, where $x$ indexes over all edges in the input graph. 
First, we compute the attention score for a message as in~\cite{vaswani2017attention}:
\begin{align}
    s^{(h)}_{\boldsymbol{\mu}_x} &= \vec{q}^{(h, r)}_i \cdot \vec{k}^{(h, r)}_j \enspace ,
\end{align}
where $\boldsymbol{\mu}_x$ is the message sent along an edge $j \rightarrow i$ labeled by the relation $r$.
Please note that a node $i$ may receive multiple messages from a node $j$, and that those messages could contain the same relation.

The scores are normalized over all messages that node $i$ receives:
\begin{align}
    \alpha^{(h)}_{\boldsymbol{\mu}_x} &= \dfrac{s^{(h)}_{\boldsymbol{\mu}_x}}{\sum_{\boldsymbol{\mu}_{x^{'}}} s^{(h)}_{\boldsymbol{\mu}_{x^{'}}}}
\end{align}

The normalized scores are used to compute a summary:
\begin{align}
    \vec{z}^{(h, k)}_i &= \sum_{\boldsymbol{\mu}_{x}} \vec{v}^{(h, r)}_j \alpha^{(h)}_{\boldsymbol{\mu}_x} \enspace ,
\end{align}
where $r$ is the relation of $\boldsymbol{\mu}_x$ and $j$ is the source node id and $i$ is the target node id of message $\boldsymbol{\mu}_x$.

The updated representation for node $i$ is then the concatenation over all heads, fed through the activation function $\sigma$:
\begin{align}
    \vec{z}^{(k)}_i &= \sigma([\vec{z}^{(0, k)}_i, \dots, \vec{z}^{(H, k)}_i]) \enspace .
\end{align}

We experiment with a ReLU, as well as with a linear $\sigma$.

\subsection{Conditional Recall}
\label{sec:recall}
In this experiment, we aim to evaluate the ability of the different GNNs to (i) remember node labels over a large number of hops and (ii) learn simple reasoning rules for sequences.

\subsubsection{Task Setup:}
We define a sequence classification task where given a sequence of characters $[x_1, \dots, x_N]$,
the model is asked to predict the correct class based on the computed representation of the last node (corresponding to input $x_N$).
The input sequences consist of strings of letters and numbers of a given length (which was varied between different experiments).
The class of the sequence is determined by the following rules:
(i) if there is a digit in the sequence, the first digit corresponds to the class label;
(ii) otherwise, if there is an upper case character, the first upper case character is the class label;
(iii) otherwise, the class is given by the first character in the sequence.
Some examples are: ''\textbf{a}bcdefg`` $\rightarrow$ ''a``, ''abc\textbf{D}efg`` $\rightarrow$ ''D``, ''abcd\textbf{3}Fg`` $\rightarrow$ ''3``, ''abCd\textbf{3}fg`` $\rightarrow$ ''3``.
We generated 20 examples per output class for a total of 1220 examples and split the data in 80/10/10 train/validation/test splits.



We transform the input sequences into a graph by 
(i) creating a node for every character of the sequence and 
(ii) adding edges \texttt{next} and \texttt{previous} between every adjacent element in the sequence. 
We also add self-edges from each node to itself.
Given these edges, the GNN has to use at least $N$ layers/steps in order to propagate information from the first node to the last node.
Readout for prediction is done by taking the representation of the last node.
\begin{figure}
    \centering
    \includegraphics[scale=0.55]{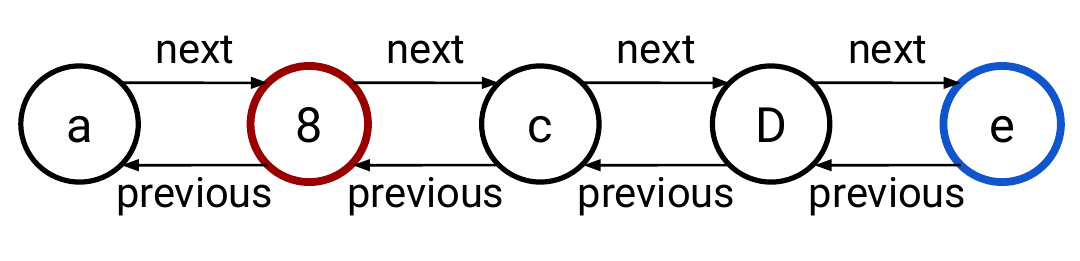}
    \caption{Example of input graph for the Conditional Recall task. The top-level state of the blue node is used for prediction. The red node specifies the desired output.}
    \label{fig:recall:inputgraph}
\end{figure}
An example of the conditional recall task and its graph representation are given in Figure ~\ref{fig:recall:inputgraph}.
In this example, the input sequence is ``\textit{a8cDe}''.
Each of the characters gets its own node and the nodes are connected using \texttt{next} and \texttt{previous} edges.
According to the rules, the answer for this input is ``\textit{8}''.
The node corresponding to the output class is highlighted in red in the figure.
The readout happens at the node corresponding to the last element in the input sequence, in this case the node with label ``\textit{e}'', which is marked in blue in the figure.

While a simple GRU or LSTM would solve the task well, and different graph representation and readout \footnote{For example, we could add an (artificial) ``master'' node that is connected to every node in the original graph, which is a common trick used in graph tasks to bypass the long-term reasoning limitations of baseline GNNs.
Also, readout by pooling over all nodes is commonly used as well.} methods
would perform better, here we deliberately choose the above formulation to measure how well the different GNNs can learn over a larger numbers of hops in a simplest graph with large diameter.

\subsubsection{Experimental details:}
\label{app:exp1}
For our method, SGGNN-RVGAT, in our experiments, we vary the dimensionality of the node states, depending on the length $l$ of the input sequence of the task ($l < 10$ $\rightarrow$ $D$ = 100, $l = 10$ $\rightarrow$ $D$ = 120, $l > 10$ $\rightarrow$ $D$ = 200).
For our baselines, we run hyperparameter search for different values of $l$, varying node state $D$ between 100 and 200, dropout rate between 0 and .5. We generally keep our batch size at 20, except for higher  numbers of examples per class, where we set batch size to 50.
Throughout most of experiments, unless otherwise indicated, we use the Adam~\cite{adam} optimizer with a learning rate of 0.001. We also use label smoothing with a factor 0.1.

The initial node states are initialized by embedding the node type using a low-dimensional embedding matrix (dimensionality 20; for a vocabulary of 62 characters) and projecting the low-dimensional embeddings to the node state dimension.

\subsubsection{Results and Discussion:}

\begin{table}[t]
    \centering
    \begin{tabular}{l|c c c c}
    \toprule
                     & 3    & 5    & 7    & 10\\
     \midrule
    RGCN             & 96.7 & 78.7 & 38.5 & 6.6 \\
    GGNN             & 96.7 & 74.6 & 17.2 & 2.5 \\
    RGAT            & 96.7 & 90.2 & 0.0 & 0.0 \\
    \midrule
SGGNN-RV-GAT          & 98.4 & 96.7 & \textbf{91.8} & \textbf{96.7} \\
\midrule
GGNN-RV-GAT           & \textbf{100.0} & 18.9 & 13.1 & 4.1 \\
SGGNN-RV-\textit{mean}               & 95.9 & 94.3 & 33.6 & 11.5\\
SGGNN-RM-GAT          & 98.4 & \textbf{98.4} & 84.4 & 8.2 \\
    \bottomrule
    \end{tabular}
    \caption{Conditional recall results for shorter sequences.
    }
    \label{tab:recall:short}
\end{table}

We show performance of the different GNNs for different 
sequence lengths in Table~\ref{tab:recall:short}.
The proposed model retains a high accuracy 
for longer sequences, demonstrating the ability to (i) learn simple rules over a large number of hops in the graph and (ii) contain a large number of layers (e.g., we employed a GNN with depth 31 for sequences of length 30).

The results for RGCN, GGNN, and RGAT in the first half of Table~\ref{tab:recall:short} reveal that these previously proposed methods are able to solve the task to a satisfactory degree only up to length 5.
For task lengths 7, their performance degrades drastically and falls under 10\% for task length 10. 
\dl{Note that even though the GGNN uses a GRU, as explained before, the GRU is applied ``vertically'', and only prevents vanishing gradients in the depth of the network (towards the initial states), not in the width (towards the neighbours).}

The results for different ablations of our model in the bottom half of Table~\ref{tab:recall:short} show that the three components are necessary to achieve the best performance, with the change from SGRU to GRU resulting in the worst performance decrease.




\subsection{Tree Max}
\label{sec:treemax}
In this experiment, we aim to evaluate the ability of a GNN to retrieve node labels over a large number of hops for many nodes simultaneously.

\subsubsection{Task Setup:}
We define a node classification task on tree-shaped graphs as described in the following.
The input graphs are trees with nodes labeled with random integers between 1 and 100.
The target labels for node classification are defined as the largest value of all the descendants of a node and the node itself.
The graphs contain edges from a parent to its children, and from children to their parent, as well as self-edges.
We use numbered child edges and numbered child-of edges,
for example \textsf{:CHILD-1-OF} for the edge going from the first child of a node to its parent.

\begin{figure}[]
    \centering
    \includegraphics[scale=0.45]{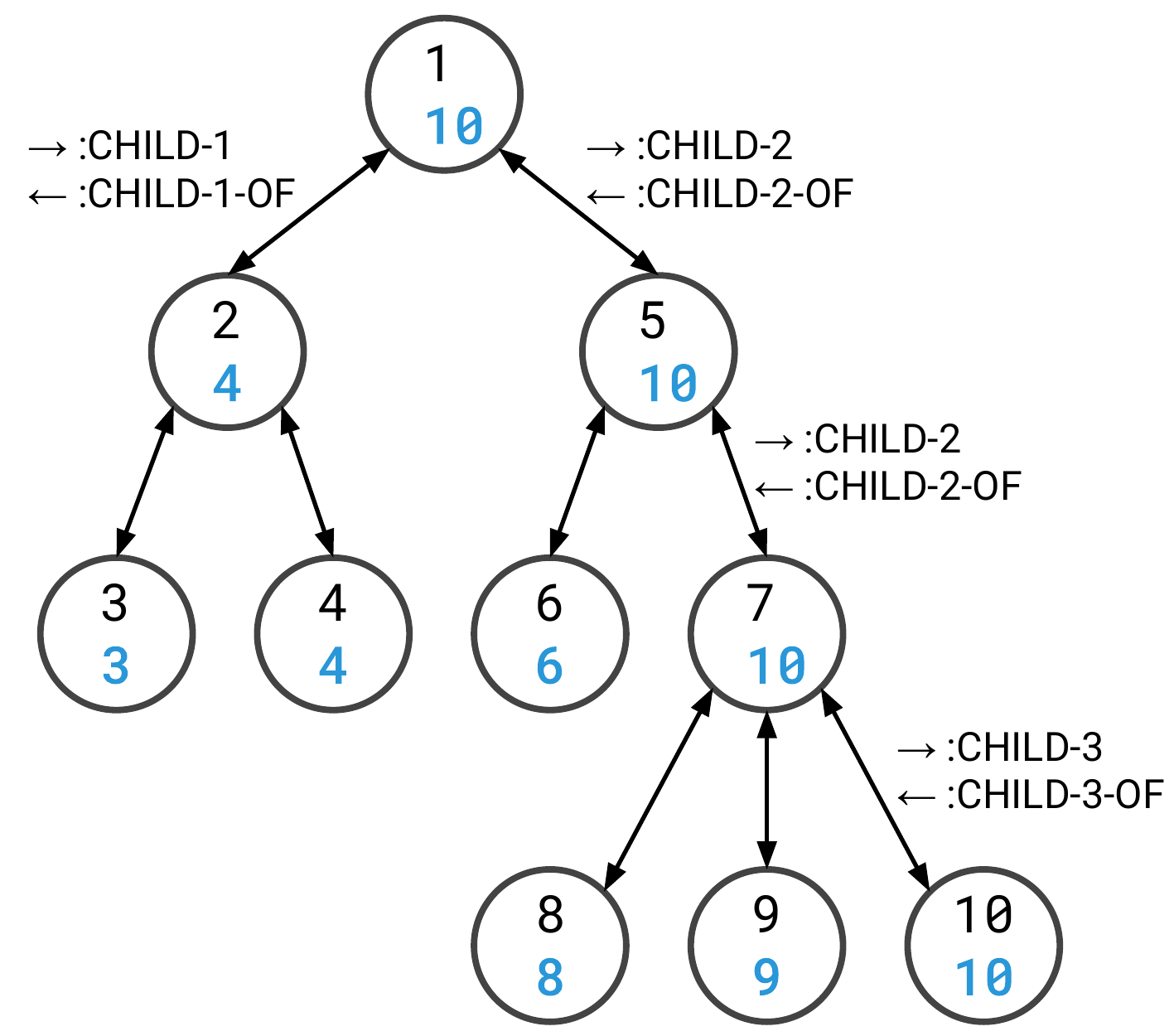}
    \caption{Example of input graph for the Tree Max task. 
    The double-ended arrow between a node $v$ and a node $u$ represents two edges: one going from $v$ to $u$ and the other from $u$ to $v$.
    Not every arrow is labeled for clearer presentation.
    Self-edges are omitted for clearer presentation as well.
    The labels on the arrows indicate edge labels: the forward arrow $\rightarrow$ corresponds to the label of the edge going from the parent to the child and $\leftarrow$ to the label of the edge going from the child to the parent.
    The input labels are the black numbers in the node circles and the output labels are the blue numbers in the node circles.
    }
    \label{fig:treemax:graph}
\end{figure}

As an example of this task, consider the following input tree:
\begin{center}
\textit{(1 (2 (3 ) (4 )) (5 (6 ) (7 (8 ) (9 ) (10 ))))}
\end{center}
This tree has a depth of four.
The nodes in the tree should be labeled with the value of their highest valued descendant.
Thus, the tree with its output labels will become:
\begin{center}
\textit{(10 (4 (3 ) (4 )) (10 (6 ) (10 (8 ) (9 ) (10 ))))}
\end{center}
In this case, to correctly predict the output label of the root node (1), the GNN must handle 3 hops.
In the case where the tree would have been labeled differently, and the maximum value wouldn't be in the deepest leaves, the GNN would still have to be able to handle 3 hops to ensure that the deepest leaves are not larger.
However, it would also not be noticeable if the GNN can't reach the deepest leaves from the root since the prediction can be done with less than 3 hops.
See Figure~\ref{fig:treemax:graph} for an illustration of the graph representation of this example.

During data generation, we first randomly pick a tree depth between 5 and 15.
Then, we generate a tree, choosing between 0, 2 or 3 children for each node until we reach the chosen tree depth.
The largest trees in our generated data contained over 200 nodes.
The generated dataset contained 800 examples; we used a 50/25/25 train/validation/test split.

For every model tested, we perform a hyperparameter search using a fixed random seed. 
For all models except our RGAT baseline, we used 17 layers.
Then we run the best hyperparameter setting with 5 different random seeds and report the test results in Table~\ref{tab:treemax}.
\footnote{Note that the same 5 seeds are re-used for experiments for all models, and that every seed results in a different dataset being generated.}
We used early stopping and reloaded the best model.
For more details,
please see Appendix~\ref{app:exp2}. 

We evaluated the trained models based on node-wise and graph-wise accuracy\footnote{The graph-level accuracy is 100\% for an example only if all nodes in the graph have been classified correctly, and is 0\% otherwise}. 

\subsubsection{Experimental details:}
\label{app:exp2}

We generate a dataset of 800 examples and split it into training, validation and test sets using 50\%, 25\% and 25\% of the data.

We use a 17-layer network for all examples unless otherwise indicated.
Note that in our experiments we use early stopping based on node-level classification accuracy on the validation set with patience set to 10 and the minimum number of training epochs set to 20.
We reload the best model weights after training finishes and evaluate on the test set.

In our experiments, we randomly explore different combinations of different hyperparameter settings: dropout is selected from $\{0., 0.1, 0.25, 0.5\}$, dimensionality of node vectors is selected from $\{150, 300\}$. 
Learning rate is chosen from $\{0.00075, 0.0005\}$. 
Since our SGGNN-RM-GAT ablation model showed unstable training behaviour, we used a learning rate of 0.00025.

For RGCN, we explore the following numbers of layers/steps: [4, 7, 12, 17].
For our RGAT baseline, we use [10, 14, 17].
Note that when the number of layers/steps is lower than 14, we can not achieve 100\% accuracy on the task.

We take the best hyperparameter setting for every model and run each with five different manually pre-defined random seeds. The same seeds are re-used across different models.

\subsubsection{Results and Discussion:}
\begin{table}[t]
    \centering
    \begin{tabular}{l|c c}
    \toprule
                     & node & entire graph\\
     \midrule
    RGCN             & $95.7 \pm 0.4$ & $31.7 \pm 1.5$ \\
    GGNN             & $97.0 \pm 0.2$ & $40.6 \pm 3.5$\\
    RGAT            & $88.3 \pm 1.5$ & $10.0 \pm 3.1$ \\
    \midrule
SGGNN-RV-GAT          & $\mathbf{99.7 \pm 0.1}$ & $\mathbf{87.8 \pm 3.2}$\\
\midrule
GGNN-RV-GAT           & $95.0 \pm 6.3$ & $48.1 \pm 25.$ \\
SGGNN-RV-\textit{mean} & $99.0 \pm 0.1$ & $68.5 \pm 3.2$ \\
SGGNN-RM-GAT          & $96.8 \pm 1.9$ & $45.1 \pm 18.5$\\
    \bottomrule
    \end{tabular}
    \caption{Node and graph-level accuracies over the test set of the Tree Max task. }
    \label{tab:treemax}
\end{table}


The results reported in Table~\ref{tab:treemax} show that RGCN and GGNN reach a high accuracy for node classification. 
For RGAT, we obtained the best results with 5 layers, which is insufficient to capture the longest necessary dependencies. 
We also quickly ran into memory problems, which required using gradient accumulation and made training much slower than the other models.

The best node-level and graph-level accuracies were obtained using the proposed SGGNN-RV-GAT model.
The ablation study indicates 
all components are essential for achieving the best performance.

Note that for correct node classification, most nodes don't require long-range information propagation.
In fact, only $0.421 \pm 0.035$~\% of all nodes across all examples in the test portion of our dataset require handling 10 hops or more and only $5.633 \pm 0.131$~\% require handling 5 hops or more.
\dl{So even though the graph-level accuracy may exaggerate the node-level errors, we believe both should be considered, in particular regarding long-range performance.
}


\section{Discussion and Conclusion}
\label{sec:discussion}
In this work, we proposed a novel GNN architecture that incorporates three changes into the GGNN to improve the learning of long-term dependencies in multi-relational graphs.
We perform experiments on two synthetic tasks and show that the proposed architecture outperforms several popular GNN variants on these tasks.
The proposed model beats the baselines by a significant margin and is also more parameter and memory efficient, but can be more computationally expensive because of the use of attention and gated updates.
The ablation study shows that all three changes of the proposed model w.r.t.~the GNN are essential. 

We plan to run further experiments on real datasets to ensure the transferabilty of our observations to the real world setting.
Major improvements from using our proposed model instead of the baselines only start to appear with a larger number of layers.
Even though this might not be significantly advantageous for all real tasks, we expect our approach to improve upon existing architectures for tasks that require to learn long-term dependencies, especially in more sparsely connected
graphs.
This may include tasks that use parse trees of natural language sentences, code syntax trees, or knowledge graphs.

\bibliographystyle{unsrt}  
\bibliography{references}

\end{document}